\documentclass{article}

\usepackage{microtype}
\usepackage{graphicx}
\usepackage{subcaption}
\usepackage{booktabs} 

\usepackage{hyperref}

\usepackage[accepted]{icml2026}

\usepackage{amsmath}
\usepackage{amssymb}
\usepackage{mathtools}
\usepackage{amsthm}

\usepackage[capitalize,noabbrev]{cleveref}

\theoremstyle{plain}

\theoremstyle{definition}

\theoremstyle{remark}

\usepackage[textsize=tiny]{todonotes}

\icmltitlerunning{Dimensionality Controls When Modularity Helps in Continual Learning}

\begin{document}

\twocolumn[
 \icmltitle{Dimensionality Controls When Modularity Helps in Continual Learning}



  \icmlsetsymbol{equal}{*}

  \begin{icmlauthorlist}
    \icmlauthor{Kathrin Korte}{itu}
    \icmlauthor{Christian Medeiros Adriano}{hpi}
    \icmlauthor{Joachim Winther Pedersen }{itu}
    \icmlauthor{Eleni Nisioti}{itu}
    \icmlauthor{Sebastian Risi}{itu}
  \end{icmlauthorlist}

  \icmlaffiliation{itu}{IT University of Copenhagen, Denmark}
   \icmlaffiliation{hpi}{Hasso Plattner Institute, University of Potsdam, Germany}

  \icmlcorrespondingauthor{Kathrin Korte}{kort@itu.dk}
  \icmlcorrespondingauthor{Christian Medeiros Adriano}{christian.adriano@hpi.de}

  \icmlkeywords{Continual learning, Catastrophic forgetting, Compositionality, Representational dimensionality, Modular recurrent networks, Stability–plasticity dilemma}

  \vskip 0.3in
]



\printAffiliationsAndNotice{}  

\begin{abstract}
    Compositional learning systems must balance plasticity, the ability to acquire new knowledge, with stability, the preservation of previously learned components, especially when tasks share structure and risk interference. We study how modular architecture, task similarity, and representational dimensionality jointly shape compositional continual learning in a sequential A–B–A paradigm, comparing a task‑partitioned recurrent network to a single‑network baseline while inducing high‑ and low‑dimensional regimes via weight‑scale manipulations. In a high‑dimensional “lazy” regime, both architectures achieve similar performance and internal geometry, suggesting that explicit modular structure has little impact when representations are weakly constrained. In a lower‑dimensional “rich” regime, modularity becomes decisive: the modular network develops graded task‑specific subspaces that overlap for similar tasks, partially align for moderately dissimilar tasks, and separate for dissimilar tasks, yielding a more compositional and interpretable organization than the single network. These findings identify the representational regime induced by initialization scale, which co-varies with representational dimensionality, as a key factor governing when compositional, modular structure is functionally beneficial in continual learning, and support viewing safety and robustness as problems of adaptive allocation of representational subspaces rather than fixed separation versus sharing.
\end{abstract}

\section{Introduction}

Compositionality --- the ability to construct and reuse structured representations across
contexts --- is widely regarded as a hallmark of robust generalization in both biological
and artificial systems~\cite{salatiello2026modularitybedrocknaturalartificial}. A system
that truly leverages compositional structure should be able to reuse components learned
from one task when solving related ones, while keeping sufficiently separate representations
for dissimilar tasks. In sequential learning settings, this ideal is challenged by a
fundamental tension between plasticity, to acquire new knowledge, and stability, to preserve
previously learned representations --- the so-called stability--plasticity
dilemma~\cite{holton2025humans}. Continual learning is a central challenge for artificial
and biological intelligence: an adaptive system must learn from a stream of experiences,
integrate new information, and remain effective on previously encountered
tasks~\cite{ramasesh2020anatomy}. In practice, this tension is reflected in how existing
representations are reused: shared structure can support compositional transfer across
tasks, but may also lead to interference when new learning disrupts prior
knowledge~\cite{lee2021continual, liu2023connectivity, mathis2025leveraging}.
In this sense, continual learning is not only a problem of memory retention, but also
a problem of how learned structure should be reorganized in response to changing task
demands. Learning depends not only on what is learned, but also on how experience is
structured and represented over time~\cite{menghi2025impact}.

A key factor in this trade-off is task similarity~\cite{menghi2025effects}. When
successive tasks are highly related, reusing existing representations is often advantageous
because it enables compositional transfer and faster adaptation. When tasks are dissimilar,
however, the same reuse may become harmful because it increases
interference~\cite{hiratani2024disentangling}. This creates a non-trivial decision problem:
should a learner continue to adapt an existing representation, or allocate new
representational resources and keep tasks more separate? In the language of psychology,
this resembles the tension between lumping and splitting --- learners may either consolidate
related experiences into a shared structure or carve out distinct representations when task
demands diverge~\cite{holton2025humans}. Representational geometry provides a natural
framework here, asking whether task representations are encoded in overlapping, partially
aligned, or approximately orthogonal subspaces~\cite{kriegeskorte2013representational} ---
and directly connecting compositionality to questions about how structure is shared or
separated across tasks. The optimal strategy depends on the current task, the similarity
to prior experience, and the learner's expectations about future tasks~\cite{holton2025humans}.

Several mechanisms have been proposed to mitigate interference in sequential 
learning, including replay, synaptic consolidation~\cite{kirkpatrick2017overcoming, 
rolnick2019experience}, and architectural approaches that use modular 
separation~\cite{ellefsen2015neural, salatiello2026modularitybedrocknaturalartificial}. 
Modularity is particularly relevant for compositional learning, as it can isolate 
task-specific computations while preserving the possibility of reuse across related 
tasks. Yet complete structural separation may eliminate transfer alongside 
interference~\cite{holton2025humans}, raising a deeper question: when and why does 
modularity help or hinder the flexible reuse of learned representations?

One promising answer lies in representational dimensionality. Prior work suggests 
that rich versus lazy learning regimes induce markedly different representational 
geometries~\cite{flesch2021rich, yu2025dimensionality}, and that modular organization 
tends to emerge most clearly when representations are low-dimensional and 
structured~\cite{Johnston2024.09.30.615925, johnston2024modular}. This points to a 
broader possibility: structural bias may only shape continual learning behavior when 
the representational space is sufficiently constrained for geometry to become a 
binding factor, 
suggesting that dimensionality may be one important aspect of the representational regime 
that determines when modular, compositional structure becomes functionally beneficial.

\begin{figure*}[!t]
            \centering
            \includegraphics[width=\textwidth]{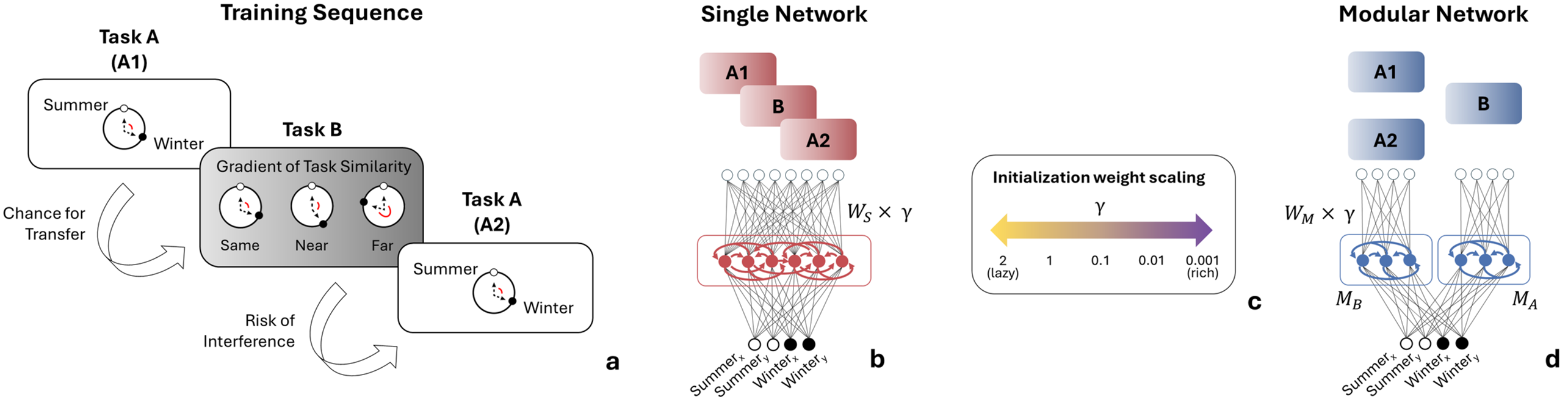}
            \caption{\textbf{Overview of the continual-learning setup, architectures, and representational regimes.}
\textbf{(a)} Sequential training protocol. Networks are trained on task A in phase $A1$, then on task B in phase $B$, and finally retested on task A in phase $A2$. Task B is instantiated in three similarity conditions relative to task A (same, near, far), highlighting a tension between the opportunity for transfer when tasks are similar and the risk of interference when learning task B alters representations used for task A.
\textbf{(b)} Single-network baseline. All inputs are processed through one recurrent population, and predictions are produced from a shared readout.
\textbf{(c)} Weight scaling. After initialization, all trainable weights (Single: $W_S$, Modular: $W_M$) are rescaled by a factor $\gamma$, which we use to span a high-dimensional “lazy’’ regime (large $\gamma$) and a low-dimensional “rich’’ regime (small $\gamma$).
\textbf{(d)} Task-partitioned modular network. Inputs are routed to task-specific recurrent modules, and module outputs are combined by a shared readout. This architecture enforces structural separation of recurrent processing, allowing us to test when such modularity alters learning dynamics and representational geometry.}            \label{fig:overview}
        \end{figure*}

In this paper, we examine this hypothesis using a sequential transfer--interference
paradigm inspired by prior work on task similarity and representational
separation~\cite{holton2025humans} (\Cref{fig:overview}a). We compare a
task-partitioned modular recurrent network (\Cref{fig:overview}d) with a
single-module baseline (\Cref{fig:overview}b), and systematically vary both task
similarity (same, near, far) and initialization scale (\Cref{fig:overview}c).
As in~\cite{holton2025humans}, we treat initialization scale as a control variable that
changes the effective dimensionality of learned representations~\cite{flesch2021rich}, and
we ask whether architecturally induced structural bias only becomes meaningful in continual
learning once the representational regime becomes sufficiently constrained. To address
this, we analyze behavioral outcomes such as accuracy, transfer, and interference, as well
as the geometry of hidden-state representations, using effective dimensionality, principal
angles, and qualitative 3D PCA visualizations of task-specific trajectories.

Our main contributions are threefold. \textbf{First}, we show that the benefit of modular
architecture in continual learning is not universal but conditional: modular and
single-network baselines perform comparably in high-dimensional, lazy regimes, and diverge
only when representational dimensionality is sufficiently reduced. 
This identifies the $\gamma$-induced representational regime, which co-varies with \emph{representational dimensionality}, as a key variable that gates when structural modularity becomes functionally meaningful. \textbf{Second}, we demonstrate that in
lower-dimensional regimes, modular networks support a graded, \emph{compositional}
organization of representational subspaces: similar tasks remain aligned, moderately
dissimilar tasks partially separate, and dissimilar tasks strongly orthogonalize --- a
structured geometry that is absent in the single-network baseline. \textbf{Third}, we
connect these findings to a geometric view of adaptive representational allocation,
arguing that the goal of continual learning should not be maximal task separation, but
\emph{similarity-dependent geometry} --- overlapping representations when tasks share
structure, partial reorganization when they are moderately related, and stronger separation
when they diverge. Together, these results suggest that robust compositional continual
learning requires mechanisms that regulate both representational dimensionality and
subspace organization dynamically as task structure changes.

\section{Background and Related Work}

\subsection{Continual learning, stability–plasticity, and task similarity}

Continual learning studies how a system can acquire new tasks without catastrophically forgetting previous ones, a problem often framed as the stability–plasticity dilemma: preserving past performance requires constraining changes to established representations, whereas adapting to new tasks benefits from reusing and reshaping those same representations \cite{lee2021continual,holton2025humans}. Methods such as regularization, replay, and rehearsal mitigate forgetting by either constraining important parameters or re-presenting past data during training \cite{kirkpatrick2017overcoming,robins1995catastrophic,rolnick2019experience,wickramasinghe2023continual,van2020brain}. A central insight from this literature is that whether reuse is beneficial depends strongly on task similarity: related tasks can support forward transfer, while dissimilar tasks make shared representations a source of interference and overwrite \cite{lee2021continual,menghi2025impact,holton2025humans,wakhloo2026neural}. Representational geometry provides a natural language here by asking whether tasks occupy overlapping, partially aligned, or approximately orthogonal subspaces in hidden-state space \cite{flesch2022representation}. This perspective shifts attention from performance alone to the \emph{format} of internal representations, and it shows that rich versus lazy learning regimes induce different geometries and different trade-offs between speed, robustness, and interference \cite{flesch2022representation,seguin2022network}. In sequential A–B–A paradigms, human data suggest that low-dimensional “rich” regimes can support similarity-dependent organization: overlapping codes for similar tasks, partial separation for moderately related tasks, and more orthogonal organization when tasks are dissimilar \cite{holton2025humans,menghi2025impact}.

\subsection{Dimensionality in biological and artificial circuits}

Neural population activity in biological circuits often lies on low-dimensional manifolds that capture task-relevant latent variables, but the link between dimensionality and performance is heterogeneous across areas and tasks \cite{yu2025dimensionality}. Compressed codes can support stronger generalization and robustness in some settings, whereas higher-dimensional representations preserve detailed information and can facilitate flexible reuse in others \cite{yu2025dimensionality}. In deep networks, effective dimensionality depends on both architecture and optimization: “rich” regimes tend to learn structured, often lower-dimensional internal codes, while “lazy” regimes keep representations closer to their initialization and thus higher-dimensional and less specialized \cite{flesch2022representation}. Prior work shows that these regimes differ in how they support transfer and interference across tasks, especially when tasks share structure only partially \cite{flesch2022representation,holton2025humans}. For continual learning, this suggests that the usefulness of any architectural bias—such as modularity—may depend on whether representational capacity is sufficiently constrained for geometry to become a binding factor, rather than on architecture alone.

\subsection{Modularity as architectural prior and constraint}

Modularity has been proposed as a structural prior for reducing interference in sequential tasks by allocating distinct subsets of parameters to different computations \cite{ellefsen2015neural,mathis2025leveraging,achterberg2023spatially,salatiello2026modularitybedrocknaturalartificial}. In artificial systems, modular architectures can emerge under evolutionary or optimization pressures and often support more robust behavior when tasks compete for limited representational resources \cite{clune2013evolutionary,yue2017brain,ellefsen2015neural}. In neuroscience, modular or partially specialized subpopulations have been linked to task complexity, connectivity constraints, and physical organization \cite{yue2017brain,gu2024emergence}. Recent work shows that specialization within modular recurrent systems can be dynamic: modules may de-specialize when information is globally accessible and re-specialize when access is constrained, even when coordination occurs only through a shared readout \cite{bena2025dynamics}. This supports a view in which modularity is not a fixed property but an emergent response to structural constraints on information flow and resource allocation.

The impact of modular architecture, however, appears to depend on representational dimensionality. When inputs or hidden representations are high-dimensional, even complex tasks may be solvable by a simple readout without requiring explicit modular separation, and learned codes can remain relatively unstructured or only implicitly modular \cite{Johnston2024.09.30.615925}. In contrast, explicit modular organization tends to emerge and become functionally meaningful when inputs or hidden states are effectively low-dimensional, forcing the network to organize computations into more clearly separated subspaces \cite{Johnston2024.09.30.615925,yu2025dimensionality,lu2025rethinking}. Under fixed parameter budgets, different architectures realize different balances between stability and plasticity, suggesting that the stability–plasticity trade-off is partly architectural rather than purely algorithmic \cite{lu2025rethinking}. Taken together, this work motivates the central question of the present study: \emph{when} and under \emph{which} representational regimes does structural modularity actually reshape representational geometry and continual learning behavior, beyond what a single-network architecture would achieve?

\section{Methods}
        \subsection{Experimental Design}
        
        We studied continual learning in a sequential transfer-interference task originally introduced by Holton et al.~\cite{holton2025humans}. We use this task only as a controlled experimental paradigm for studying continual learning in neural networks, not as a model of human behavior. The task consists of three phases: \(A1 \rightarrow B \rightarrow A2\) (\Cref{fig:overview}a). In phase \(A1\), the network learns task A. In phase \(B\), it learns task B on a new set of stimuli. In phase \(A2\), task A is revisited to quantify retention and interference after exposure to task B (\Cref{fig:overview}a). Each task requires mapping six discrete plant cues to positions on a circular dial. For each plant, locations in the two seasonal contexts, \emph{summer} and \emph{winter}, are related by a fixed angular offset, referred to as the \emph{task rule}. Knowing a plant's location in one season and the task rule determines its location in the other season. Across schedules, the numerical value of the rule is randomized. Task B uses novel stimuli but the same formal structure as task A. The \emph{Same}, \emph{Near}, and \emph{Far} conditions manipulate how the task B rule relates to the task A rule: the rules are identical in the Same condition, shifted by a small angle in the Near condition, and shifted by a large angle in the Far condition. All other design features are held fixed across these conditions~\cite{holton2025humans}. Recurrent networks were trained on trial sequences derived from the original experiment. These sequences matched the original task structure in trial order, stimulus identity, probed season, feedback availability, and task rule. The networks were trained independently and were not fit to human motor responses. Each stimulus was encoded as a one-hot input vector. The network produced a four-dimensional output consisting of two cosine-sine pairs: the first pair encoded the summer location (\texttt{feature\_idx}=0), and the second pair encoded the winter location (\texttt{feature\_idx}=1). On each trial, mean squared error (MSE) was calculated only for the output pair corresponding to the probed season, while the other pair did not contribute to the loss for that step. Angular accuracy was computed by reconstructing the predicted angle from the supervised cosine-sine pair and comparing it to the target angle on the circle.
        
        Training was organized in three phases: \(A1\) (task A acquisition), \(B\) (task B acquisition on new stimuli), and \(A2\) (retest on task A). Each phase followed the same trial-wise alternation between summer and winter probes as in the original design. Following Holton et al., transfer was quantified using winter-trial accuracy: the change from the end of task A to the beginning of task B, operationalized in our analysis pipeline as the mean over the last six winter trials in \(A1\) versus the first six winter trials in \(B\). Interference at task A retest was quantified from winter responses in phase \(A2\) using the mixture-based measure described in the Analysis section, analogous to Holton et al.'s probability-of-updating-to-the-task-B-rule treatment. Where we report generalization within task A, this refers to winter performance on held-out stimuli following the paper-faithful testing protocol when applicable. For more details, see \Cref{sec:appendix:experimental-details} in the Appendix.
        
        \subsection{Architectures}
        
        We compared two recurrent architectures: a single-network baseline and a task-partitioned modular network. The single-network baseline processes all inputs through one recurrent population. In contrast, the modular architecture contains two recurrent modules. Task identity determines how inputs are routed: task A stimuli are delivered to module \(M_A\), and task B stimuli are delivered to module \(M_B\). The outputs of the modules are combined by a shared readout. Let \(x_t\) denote the input at time \(t\), \(h_t^A\) and \(h_t^B\) the hidden states of the two modules, and \(y_t\) the output. In the modular model, the input is partitioned into task-specific slices, $x_t^A = m_A \odot x_t$, $x_t^B = m_B \odot x_t$, where \(m_A\) and \(m_B\) are binary masks selecting the task-relevant input dimensions. Each module evolves according to
        \[
        h_t^A = \tanh\!\big(W_{\mathrm{ih}}^A x_t^A + W_{\mathrm{hh}}^A h_{t-1}^A\big),
        \]
        \[
        h_t^B = \tanh\!\big(W_{\mathrm{ih}}^B x_t^B + W_{\mathrm{hh}}^B h_{t-1}^B\big),
        \]
        with recurrent processing occurring separately in each module and no inter-module recurrent communication in the main analysis. The pre-readout state is the concatenation of the two module states
        and the shared readout maps this state to four outputs (two cosine-sine pairs):
        \[
        y_t = W_{\mathrm{out}} h_t.
        \]
        Although recurrent processing is separated across modules, both modules contribute to the same output layer, such that task representations remain coupled through a shared behavioral objective.
        
        For the single-network baseline, the same recurrent update is applied to a single hidden population:        $h_t = \tanh\!\big(W_{\mathrm{ih}} x_t + W_{\mathrm{hh}} h_{t-1}\big)$, $y_t = W_{\mathrm{out}} h_t$. This design isolates the effect of task-partitioned input processing and recurrent separation while keeping the output format comparable across architectures.
        
        \subsection{Training Procedure}
        
        Each model was initialized from a new random seed and trained on a single task sequence following the $A1 \rightarrow B \rightarrow A2$ protocol. Different model runs, therefore, correspond to different trial sequences derived from the original experimental schedules with independent random initializations. To vary the representational regime, we rescaled all trainable parameters after default PyTorch initialization by a global factor $\gamma$ (\Cref{fig:overview}c). Larger values of $\gamma$ increase the magnitude of the initial weights, whereas smaller values reduce it. This manipulation is commonly associated with transitions between so-called lazy and rich learning regimes, which in turn are linked to differences in effective representational dimensionality \cite{flesch2021rich}. 
        In this work, we treat $\gamma$ as a practical probe of rich versus lazy representational regimes rather than as a pure manipulation of dimensionality. Consequently, our conclusions should be interpreted as applying to $\gamma$-induced learning regimes, which co-vary with representational dimensionality, optimization dynamics, and conditioning.
        All models were trained using MSE loss on a cosine-sine encoding of the target angle. On each trial, the loss was computed only on the output components corresponding to the currently probed feature, ensuring that learning signals matched the task structure of the sequential protocol. 
        The single-network architecture used 50 recurrent units, while the modular architecture used two isolated recurrent modules of 25 units each with a shared linear readout. Models were trained with SGD (learning rate 0.01) for 100 epochs per phase. Inputs were unrolled for two recurrent timesteps and hidden states were reset between trials, such that recurrence operated only within individual trials while weights were updated continuously across the protocol. Initialization scale was varied using ($\gamma \in {0.001, 0.01, 0.1, 1.0, 2.0}$). Results are reported across 305 participant-derived training schedules from the original experiments \cite{holton2025humans}.
        
        \subsection{Behavioral Measures}

        We quantified model performance using accuracy, transfer, and interference. Accuracy refers to angular prediction accuracy after reconstruction from the cosine-sine output. Transfer was measured as mean winter-trial accuracy in the first six trials of phase \(B\) minus mean winter-trial accuracy in the last six trials of phase \(A1\), capturing how much prior learning facilitates early performance on the new task. Interference was quantified from responses in phase \(A2\) by fitting a von Mises mixture to $A2$ winter responses and taking one minus the mixture weight on the task-A component. This measures how much behavior on task A has shifted toward task B after learning task B, and thus reflects the degree to which previously learned responses have been overwritten. We adopt the mixture-based metric from Holton et al.~\cite{holton2025humans} to distinguish shifts toward the task-B rule from generic performance degradation and to maintain comparability with the original paradigm. 
        
        \subsection{Representational Analysis}
        
        To characterize the learned hidden-state geometry, we extracted the hidden representation at the final time step for each stimulus and phase. For each phase, the hidden states were stacked into a data matrix and analyzed using principal component analysis (PCA). We used the number of principal components required to explain 99 \% of the variance as a measure of effective dimensionality. We also computed principal angles between task-specific subspaces. Using hidden states after phase \(B\) from a canonical stimulus sweep, we split the rows into two groups according to the feature associated with each stimulus under the stored ordering and fitted a PCA within each group. We then reported the largest principal angle between the resulting two-dimensional subspaces. For the qualitative geometry analysis, we projected hidden states onto the first three principal components of a single PCA fit computed jointly over the \(A1\), \(B\), and \(A2\) phases. These 3D PCA projections provide an intuitive illustration of how task representations reorganize across learning and how this reorganization depends on architecture and initial weight scale. For more details, see the \Cref{sec:appendix:analysis-details} in the Appendix .

 \begin{figure*}[t]
            \centering
            \includegraphics[width=\linewidth]{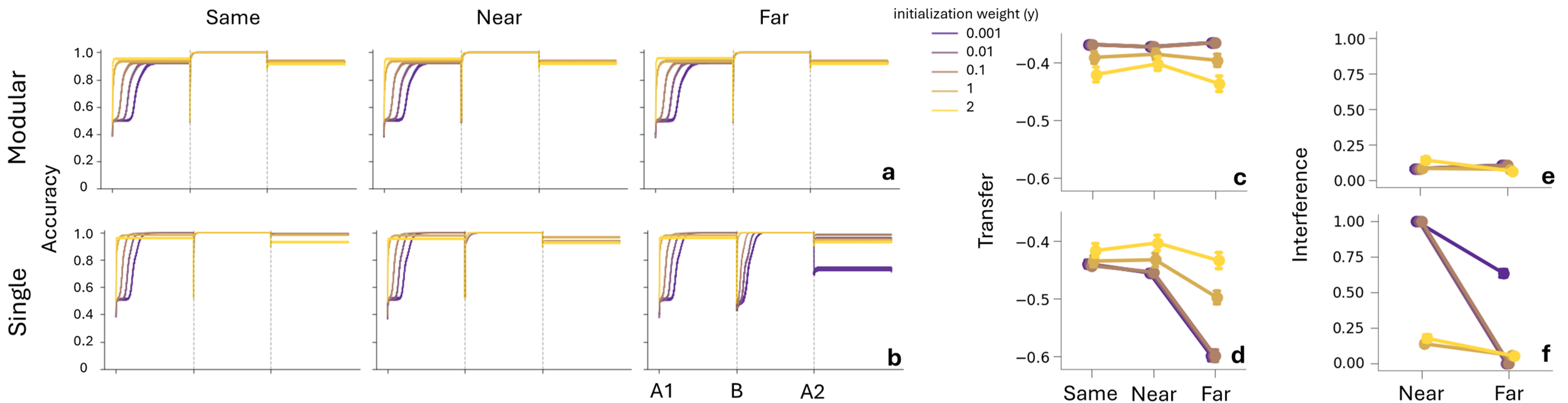}
\caption{\textbf{Modular structure attenuates transfer--interference costs in sequential learning under representational constraints.}
\textbf{(a, b)} Accuracy across the sequential $A1 \rightarrow B \rightarrow A2$ protocol for the modular network \textbf{(a)} and the single network \textbf{(b)} under the three task-similarity conditions (same, near, far) and across initialization weight scales $\gamma$. In both architectures, learning on $A1$ and $B$ rapidly reaches high accuracy, but the single network shows a stronger drop in $A2$ performance in the far condition at the smallest initialization scales, indicating greater susceptibility to interference.
\textbf{(c, d)} Transfer as a function of task similarity and initialization scale for the modular \textbf{(c)} and single \textbf{(d)} architectures. The modular network exhibits relatively stable, low transfer across similarity conditions and scales, whereas the single network becomes increasingly sensitive to task dissimilarity, with the far condition showing the strongest degradation at small $\gamma$.
\textbf{(e, f)} Interference for the modular \textbf{(e)} and single \textbf{(f)} architectures. Interference remains low in the modular network across all conditions, while the single network shows substantially larger interference, especially in the near and far conditions and at the lowest initialization scales. Together, these panels show that architectural separation reduces sequential interference, and that this benefit is most pronounced when representational dimensionality is strongly constrained by small $\gamma$.}
            \label{fig:accuperf}
        \end{figure*}

        \begin{figure}[!t]
            \centering
            \includegraphics[width=3.2in]{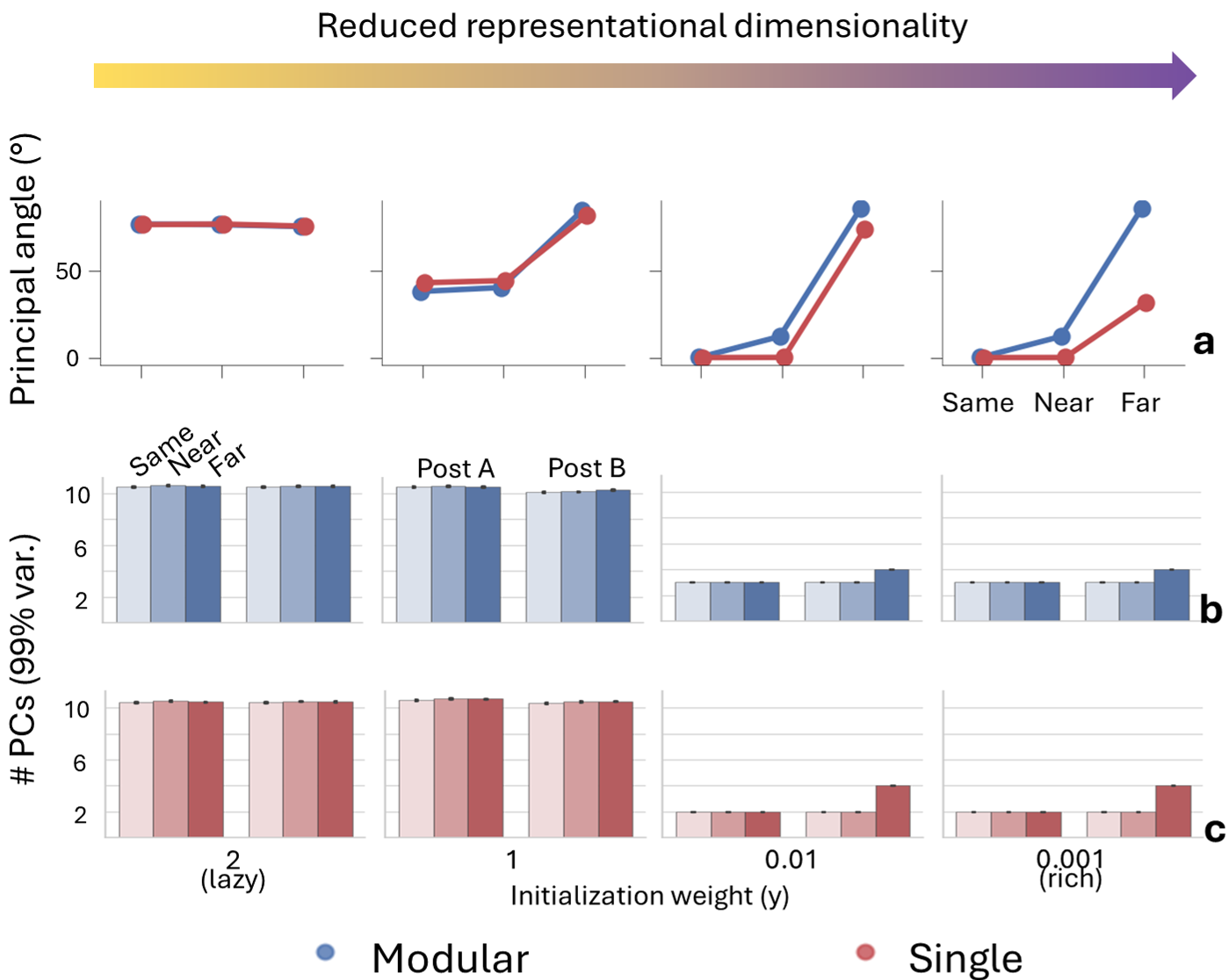}
           \caption{\textbf{Initialization weight scaling controls representational dimensionality and reveals architecture-dependent geometry.}
\textbf{(a, b)} Effective dimensionality of hidden representations, measured as the number of principal components required to explain 99\% of the variance, for modular \textbf{(a)} and single \textbf{(b)} architectures. Columns correspond to decreasing initialization weight scale $\gamma$, moving from a high-dimensional “lazy’’ regime (large $\gamma$) to a low-dimensional “rich’’ regime (small $\gamma$). Within each column, bars show task-similarity conditions (same, near, far). Dimensionality remains high at large initialization scales and decreases strongly as $\gamma$ is reduced for both architectures.
\textbf{(c)} Principal angles between task-specific representational subspaces (same, near, far) across the same initialization scales. In the high-dimensional regime, both architectures show similar geometry with relatively small differences between conditions. In the low-dimensional regime, the modular network exhibits a clearer similarity-dependent organization, with near tasks occupying intermediate angles and far tasks showing stronger separation, whereas the single network displays a weaker and less structured dependence on task similarity. Together, these results indicate that architectural separation becomes geometrically consequential only once representational dimensionality is sufficiently reduced, at which point task similarity begins to systematically shape subspace organization.}            \label{fig:reduceddim}
        \end{figure}

        \begin{figure*}[!t]
            \centering
            \includegraphics[width=\linewidth]{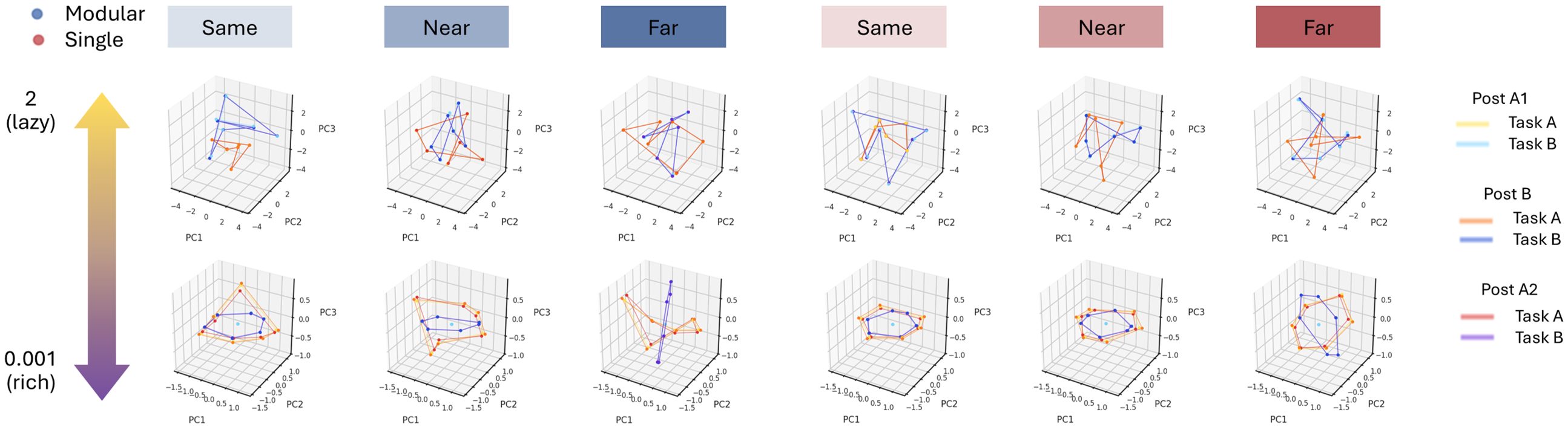}
            \caption{\textbf{3D PCA projections reveal similarity-dependent organization of task representations under reduced dimensionality.} Each panel shows hidden-state trajectories projected onto the first three principal components of a PCA fitted jointly to Post $A1$, Post $B$, and Post $A2$ activations for a given network and initialization regime. For each phase, trajectories are computed over a fixed sweep of 12 inputs, with the first six corresponding to Task A and the last six to Task B. For each phase and task, closed loops connect stimuli in sweep order. The left block shows the modular network, and the right block shows the single network. Columns correspond to task similarity conditions (same, near, far). The top row shows the lazy (high-dimensional) regime, and the bottom row shows the rich (low-dimensional) regime. Colors encode phase and task identity (Task A: warm colors, Task B: cool colors). In the high-dimensional regime (top row), trajectories occupy a broad state space and exhibit relatively weak structure, with limited differences between architectures. In the low-dimensional regime (bottom row), trajectories become more compact and their relative arrangement reflects task similarity more clearly. In the modular network, same-task trajectories remain relatively closely aligned across phases, near tasks exhibit intermediate reorganization, and far tasks show stronger separation. This similarity-dependent structure is less pronounced in the single network, which in turn exhibits closer alignment in the same regime. Overall, the figure illustrates that reduced representational dimensionality reveals a constrained geometric organization of task representations, within which architectural separation leads to more structured, similarity-dependent alignment.}
            \label{fig:reprgeom}
        \end{figure*}

\section{Results}

We first ask how architectural separation and representational regime jointly shape behavioral performance in the sequential $A1 \rightarrow B \rightarrow A2$ protocol, and then relate these effects to changes in effective dimensionality and representational geometry. Across analyses, we treat the initialization scale $\gamma$ as a practical control on the representational regime, ranging from a high-dimensional “lazy’’ regime at large $\gamma$ to a low-dimensional “rich’’ regime at small $\gamma$ (\Cref{fig:accuperf}, \Cref{fig:reduceddim}). 

\subsection{Modularity selectively reduces interference in constrained regimes}

\Cref{fig:accuperf} summarizes accuracy, transfer, and interference for the modular and single architectures across task similarity conditions and initialization scales. In both architectures, accuracy on $A1$ and $B$ rapidly approaches ceiling, confirming that all regimes can solve the constituent tasks. However, the stability of task A after learning task B differs systematically between architectures and regimes. In the modular network, $A2$ accuracy remains high across all similarity conditions and $\gamma$ values, transfer varies only modestly, and interference stays low throughout, indicating that task-partitioned input routing largely protects task A from overwrite during the $B$ phase. In contrast, the single network becomes increasingly sensitive to task dissimilarity as the regime becomes richer: while it also reaches high performance on $A1$ and $B$, $A2$ accuracy drops most strongly in the far condition at the smallest $\gamma$ values, and the interference measure increases, reflecting a larger transfer–interference cost when tasks are dissimilar and representations are constrained (\Cref{fig:accuperf}). 

The transfer and interference summaries reinforce this pattern. For the modular architecture, transfer remains relatively stable across similarity conditions and scales, and interference is consistently low. For the single network, transfer becomes less stable and interference grows with decreasing $\gamma$, especially for the near and far conditions. Taken together, these results show that architectural separation does not yield a uniform performance benefit: in high-dimensional regimes or when tasks are effectively the same, the single network performs comparably to the modular model, but in low-dimensional regimes with dissimilar tasks, modularity substantially attenuates interference while maintaining good performance on both tasks (\Cref{fig:accuperf}).

\subsection{Initialization scale controls effective dimensionality}

To test whether these behavioral differences track changes in representational regime, we quantified the effective dimensionality of hidden states using PCA on trial-averaged activity from each phase. Effective dimensionality was defined as the number of principal components required to explain 99\,\% of the variance. \Cref{fig:reduceddim}a,b shows that, for both architectures, dimensionality remains high at large $\gamma$ and decreases sharply as $\gamma$ is reduced, confirming that initialization scale serves as a practical proxy for transitioning between high- and low-dimensional regimes. At large $\gamma$ values, corresponding to the lazy regime, both architectures operate in a high-dimensional space, and dimensionality varies only weakly with task similarity. As $\gamma$ decreases, the representation collapses into a lower-dimensional manifold, and differences between architectures become more pronounced (\Cref{fig:reduceddim}a,b).

These results indicate that the behavioral effects in \Cref{fig:accuperf} co-occur with a transition between representational regimes rather than arising from architecture alone. When dimensionality is high, both architectures have sufficient degrees of freedom to encode tasks A and B with limited geometric pressure, and the benefit of structural separation remains small. When dimensionality is reduced, however, representational capacity becomes a binding constraint: tasks must compete for a limited set of directions in state space, and architectural bias starts to shape how these directions are allocated. 
In this regime, the modular architecture gains a behavioral advantage, consistent with the idea that modularity becomes most consequential in the low-dimensional representational regime induced by small $\gamma$ (\Cref{fig:accuperf}, \Cref{fig:reduceddim}).

\subsection{Modularity induces similarity-dependent task geometry}

We next examined how task similarity and architecture jointly shape the geometry of hidden-state subspaces. Using hidden states from phase $B$, we estimated task-specific subspaces via PCA and computed principal angles between them for the same, near, and far conditions. In the high-dimensional regime at large $\gamma$, both architectures show broadly similar geometry: principal angles vary only weakly with task similarity, and subspaces for different conditions remain relatively entangled (\Cref{fig:reduceddim}c). In this regime, structural separation has little observable impact on how task representations are arranged in state space.

In the low-dimensional regime at small $\gamma$, the modular architecture develops a clearer similarity-dependent organization. Principal angles in the modular network track the similarity manipulation: representations for the same condition remain closely aligned, near conditions occupy intermediate angles, and far conditions exhibit stronger separation, indicating that dissimilar tasks are pushed into more orthogonal subspaces (\Cref{fig:reduceddim}c). The single network also shows some sensitivity to task similarity but with a weaker and less structured dependence on condition; its task subspaces remain more entangled and do not exhibit the same graded separation pattern. These results suggest that under dimensional constraints, modularity supports a more controlled allocation of representational subspaces that respects task similarity, whereas the single architecture relies on less structured reuse.

\subsection{Low-dimensional regimes reveal structured trajectories}

To provide a qualitative view of these effects, \Cref{fig:reprgeom} shows 3D PCA projections of hidden-state trajectories across phases $A1$, $B$, and $A2$ for each architecture, similarity condition, and regime. In the high-dimensional (lazy) regime, trajectories occupy a broad region of state space, and both architectures exhibit relatively diffuse geometry with limited differences: task- and phase-specific loops overlap substantially, and the similarity manipulation is visible but not strongly constrained (\Cref{fig:reprgeom}, top row). This aligns with the dimensionality analyses, where architectural effects on geometry are weak when the representation is high-dimensional (\Cref{fig:reduceddim}a–c).

In the low-dimensional (rich) regime, trajectories collapse into a more compact geometry, and the arrangement of loops reflects task similarity more clearly. In the modular network, same-task trajectories remain closely aligned across phases, near tasks show intermediate reorganization, and far tasks peel off into more distinct directions, yielding a graded, similarity-dependent layout of task-specific loops (\Cref{fig:reprgeom}, bottom left). The single network also exhibits a reduction in dimensionality, but its trajectories remain more co-occupied and less neatly ordered by similarity; the separation between near and far conditions is weaker, and task A and B loops remain more intertwined (\Cref{fig:reprgeom}, bottom right). These qualitative patterns support the interpretation that reduced dimensionality reveals a constrained geometric regime in which architectural separation produces a more structured and interpretable organization of task representations.

\subsection{Ablations confirm robustness of architectural effects}

Finally, we evaluated whether the observed patterns depend on specific architectural choices by performing ablations over module size and input routing. Varying the size of the recurrent modules and altering the input-partitioning scheme did not change the qualitative results: in all configurations, modular architectures showed stable performance and low interference in low-dimensional regimes with dissimilar tasks, along with a clearer similarity-dependent organization of task subspaces, whereas single architectures remained more susceptible to interference and exhibited less structured geometry. These ablations indicate that the central findings—namely that representational dimensionality gates when modularity reshapes behavior and geometry—are robust to moderate architectural variations.
        
\section{Discussion}

Our results show that the benefits of architectural separation in continual learning are conditional rather than universal. Across the sequential $A1 \rightarrow B \rightarrow A2$ protocol, the modular network generally preserved performance on task A better and exhibited lower interference than the single network, but this advantage emerged primarily in lower-dimensional, ``rich'' regimes induced by small initialization scales. In high-dimensional, ``lazy'' regimes, both architectures behaved similarly: they solved tasks A and B, achieved comparable transfer, and showed only modest differences in interference and geometry. These findings support the view that representational dimensionality gates when modularity becomes functionally meaningful, rather than modularity providing a uniform advantage across regimes.

The geometric analyses clarify how dimensionality shapes this effect. When dimensionality remains high, task-specific subspaces are only weakly structured: principal angles vary little across task-similarity conditions, and 3D PCA trajectories occupy a broad region of state space for both architectures. In contrast, under reduced dimensionality, the modular network develops a graded organization of task representations: subspaces for same tasks remain closely aligned, near tasks occupy intermediate angles, and far tasks become more orthogonal, with trajectories forming more clearly separated loops across phases. The single network also reflects task similarity, but its subspaces are more entangled and its trajectories less neatly ordered, consistent with more diffuse reuse of a constrained representational space. Additional ablation experiments (\Cref{sec:appendix:ablation-studies}) varying module width, input routing, inter-module connectivity, initialization scope, and recurrent depth preserved this graded similarity–dependent geometry and the selective benefit of modularity in low-dimensional regimes. This argues against a pure capacity explanation and reinforcing our interpretation of dimensionality as a key gate on the functional impact of architecture.

An important aspect of our architecture is that separation is structural but not total. Task-specific inputs are routed to distinct recurrent modules, yet these modules share a common readout, so the model cannot simply implement two independent task systems. Instead, it must coordinate module activity at the output level, yielding a constrained division of labor rather than strict isolation. This partial coupling helps explain why we observe similarity-dependent but still overlapping representations, rather than fully segregated codes, and highlights how modularity can support compositional reuse without eliminating transfer opportunities.

These findings have several implications for continual learning. First, they suggest that the key question is not whether an architecture is modular in an absolute sense, but under which representational regimes modularity actually changes behavior and internal geometry. Second, they argue that the goal of continual learning should not be maximal separation between tasks, but \emph{similarity-dependent geometry}: overlapping representations when tasks effectively share structure, partial reorganization for related but distinct tasks, and stronger separation for dissimilar tasks. From this perspective, continual learning is best viewed as a problem of adaptive allocation of representational subspaces, rather than a binary choice between sharing and isolation.

This work has several limitations. We rely on PCA-based effective dimensionality as a proxy for representational regime, which provides only an approximate summary of intrinsic dimensionality. Initialization scale likely affects not only dimensionality but also optimization dynamics and how readily architectural biases are expressed. Our 3D PCA visualizations are qualitative illustrations rather than inferential tests. Finally, while the restricted protocol is well suited for probing transfer–interference trade-offs, it does not address longer or more heterogeneous task sequences. Consequently, it remains an open question whether the observed dimensionality-dependent effects of modularity extend to larger continual-learning benchmarks and more realistic task distributions.

Future work could extend this framework along three axes. First, applying similar analyses to richer task curricula and longer sequences would test how architectural bias and geometry evolve under repeated interference, recovery, and changing task similarity. Second, mechanisms that directly regulate dimensionality during learning—for example via regularization, bottlenecks, or structured noise—could provide a more targeted handle on representational regime than initialization alone. Third, comparing the present task-partitioned architecture to modular designs with explicit inter-module communication would clarify how different forms of structural coupling shape representational geometry and transfer–interference behavior.
\section*{Impact Statement}

This work contributes to understanding how training choices affect the structure of neural representations and the behavior of interpretability methods. There are no immediate societal risks associated with this work.




\bibliography{references}
\bibliographystyle{icml2026}

\newpage
\appendix
\onecolumn

\section{Appendix - Experimental Details}\label{sec:appendix:experimental-details}

\subsection{Task and Data Generation}

Each task followed the A1~$\rightarrow$~B~$\rightarrow$~A2 continual learning schedule adapted from \citet{holton2025humans}. Tasks consisted of 6 plant cues with angular responses on a circular dial, with summer and winter targets differing by a fixed angular offset. Inputs were 12-dimensional one-hot vectors (6 cues $\times$ 2 seasons); outputs were four continuous values (cosine--sine encodings for summer and winter targets).

\paragraph{Experimental Conditions} Experiments varied: (1) 2 architectures (modular task-routed recurrent network; single recurrent baseline), (2) 5 initialization scales $\gamma \in \{0.001, 0.01, 0.1, 1.0, 2.0\}$, and (3) 3 task similarity conditions (Same, Near, Far). Each architecture $\times$ $\gamma$ $\times$ similarity condition contained 101--103 independent runs (103 Same, 101 Near, 101 Far schedules), yielding approximately 3050 trained networks.

\paragraph{Trials per Phase} Each schedule contained 60 summer and 60 winter trials per phase (120 total). With 100 epochs per phase, this yielded 6000 summer and 6000 winter updates for A1 and B. During A2, winter trials were evaluated but did not contribute gradients (held-out probe).

\subsection{Training}

All models used SGD (learning rate: 0.01, momentum: 0.0, weight decay: 0.0, no gradient clipping) with batch size 1, fixed trial order, and 100 epochs per phase. Optimizer state was preserved across phases.

Recurrent weights used PyTorch's default uniform initialization, rescaled by $\gamma \in \{0.001, 0.01, 0.1, 1.0, 2.0\}$, defining the rich-to-lazy regime. Recurrent biases were disabled; only the readout layer had bias. Random seed was fixed at 2024; individual networks received unique weights via sequential RNG sampling.

\subsection{Behavioral Measures}

\paragraph{Accuracy} Predicted cosine--sine outputs were converted to angle $\hat{\theta}$:
\[
\text{Accuracy} = 1 - \frac{\left|\mathrm{wrap}_{[-\pi,\pi]}(\hat{\theta}-\theta)\right|}{\pi}
\]
Learning curves used a centered rolling average (window = 25 trials) averaged across runs.

\paragraph{Transfer} Performance retention across A1~$\rightarrow$~B:
\[
\text{Transfer} = \overline{\text{Acc}}_{B,\text{winter},\text{first 6}} - \overline{\text{Acc}}_{A1,\text{winter},\text{last 6}}
\]

\paragraph{Interference} Measured from A2 winter responses via two-component von Mises mixture:
\[
p(\phi) = \pi_A \cdot \mathrm{vM}(\phi;\mu_A,\kappa) + \pi_B \cdot \mathrm{vM}(\phi;\mu_B,\kappa), \quad \text{Interference} = 1 - \pi_A
\]
where $\mu_A$, $\mu_B$ are task-A and task-B rules, and $\pi_A + \pi_B = 1$. Computed only for Near and Far conditions. All metrics reported as mean $\pm$ SEM.

\subsection{Architecture Details}

\begin{table}[h]
\centering
\small
\begin{tabular}{lccccc}
\toprule
Architecture & Modules & Units/module & Total hidden & Recurrent params & Readout params \\
\midrule
Modular & 2 & 25 & 50 & 1850 & 204 \\
Single & 1 & 50 & 50 & 3100 & 204 \\
\bottomrule
\end{tabular}
\end{table}

Architectures were state-matched (equal hidden dimensionality) but not parameter-matched (single network had ~1250 additional recurrent parameters). Recurrent layers had no biases; output layer bias was enabled. In the modular architecture, no recurrent weights were shared; each module had independent recurrent and input projections with task-routed input (inactive module received zero input).


\section{Appendix - Analysis Details}
\label{sec:appendix:analysis-details}

\paragraph{Angular reconstruction and accuracy.}
On each trial, the network produces a four-dimensional output corresponding to two cosine--sine pairs, one for the summer feature (\texttt{feature\_idx} = 0) and one for the winter feature (\texttt{feature\_idx} = 1), as described in the Methods section. For a given probed feature $f \in \{\text{summer}, \text{winter}\}$, we reconstruct the predicted angle
\[
\hat{\theta}_t^{(f)} = \operatorname{atan2}\!\big(\hat{s}_t^{(f)}, \hat{c}_t^{(f)}\big),
\]
where $(\hat{c}_t^{(f)}, \hat{s}_t^{(f)})$ are the cosine and sine outputs for that feature at trial $t$, and $\operatorname{atan2}$ denotes the two-argument arctangent. The target angle is denoted $\theta_t^{(f)}$, and we compute the circular error
\[
\Delta \theta_t^{(f)} = \operatorname{wrap}\big(\hat{\theta}_t^{(f)} - \theta_t^{(f)}\big),
\]
where $\operatorname{wrap}$ maps angles to the interval $[-\pi, \pi)$. Unless stated otherwise, \emph{accuracy} is one minus the mean normalized absolute circular error,
\[
\mathrm{Acc} = 1 - \frac{1}{T} \sum_{t=1}^T \frac{\lvert \Delta \theta_t^{(f_t)} \rvert}{\pi},
\]
computed over a specified set of trials (e.g., all winter trials in a given phase). Equivalently, the per-trial accuracy $\mathrm{Acc}_t = 1 - \lvert \Delta \theta_t^{(f_t)} \rvert / \pi$ is averaged over $T$ trials. The metric is bounded in $[0, 1]$, with higher values indicating better performance.

\paragraph{Transfer measure.}
Following Holton et al.~\cite{holton2025humans}, we quantify transfer from task A to task B using performance on winter trials. Let $\mathcal{W}_{A1}^{\text{end}}$ denote the set of the last six winter trials in phase $A1$, and let $\mathcal{W}_{B}^{\text{start}}$ denote the set of the first six winter trials in phase $B$. We define
\[
\mathrm{Acc}_{A1}^{\text{end}} = \frac{1}{\lvert \mathcal{W}_{A1}^{\text{end}} \rvert} \sum_{t \in \mathcal{W}_{A1}^{\text{end}}} \mathrm{Acc}_t,
\quad
\mathrm{Acc}_{B}^{\text{start}} = \frac{1}{\lvert \mathcal{W}_{B}^{\text{start}} \rvert} \sum_{t \in \mathcal{W}_{B}^{\text{start}}} \mathrm{Acc}_t,
\]
where $\mathrm{Acc}_t$ is the single-trial accuracy defined from $\Delta \theta_t^{(f_t)}$ as above. The \emph{transfer} metric is then
\[
\mathrm{Transfer} = \mathrm{Acc}_{B}^{\text{start}} - \mathrm{Acc}_{A1}^{\text{end}},
\]
so that positive values indicate that prior learning on task A facilitates early performance on task B. In practice, transfer values are typically negative (winter accuracy drops at the $A1 \rightarrow B$ boundary because the rule has changed); less-negative values therefore correspond to better transfer between tasks.

\paragraph{Mixture-based interference measure.}
To quantify interference at retest on task A, we analyze winter responses in phase $A2$ using a mixture model over the network's implicit \emph{rule angle} on each trial. For a given run and condition, and for every winter trial $i$ in phase $A2$, we define the rule-angle residual between the network's winter and summer angle predictions on matched plant cues,
\[
\phi_i \;=\; \operatorname{wrap}\!\big( \hat{\theta}_{\text{winter},i} - \hat{\theta}_{\text{summer},i} \big),
\]
where $(\hat{\theta}_{\text{summer},i}, \hat{\theta}_{\text{winter},i})$ are reconstructed from the network's two cosine--sine output channels (output indices 0--1 and 2--3, respectively). Let $\theta_A$ and $\theta_B$ denote the task-A and task-B rule angles (the angular offsets between summer and winter targets) stored from the schedule. We fit a two-component von~Mises mixture
\[
p(\phi) = \pi_A \, \mathrm{VM}(\phi \mid \theta_A, \kappa) + (1 - \pi_A) \, \mathrm{VM}(\phi \mid \theta_B, \kappa),
\]
where $\mathrm{VM}(\cdot \mid \mu, \kappa)$ is the von~Mises distribution with mean direction $\mu$ and concentration $\kappa$. For simplicity and identifiability, we use a shared concentration parameter $\kappa$ across components and fix the component means to $(\theta_A, \theta_B)$; the only free parameters are $(\pi_A, \kappa)$, with $\pi_B = 1 - \pi_A$.

We estimate $(\pi_A, \kappa)$ by maximizing the log-likelihood
\[
\mathcal{L}(\pi_A, \kappa) = \sum_{i=1}^N \log p(\phi_i)
\]
using an expectation--maximization (EM) procedure that alternates an analytic E-step for the membership weights with an L-BFGS-B M-step on $\kappa$, run to convergence (parameter-update tolerance $10^{-3}$). To avoid local optima, we sweep a deterministic grid of starting points: $\pi_A^{(0)} \in \{0.1, 0.2, \ldots, 0.9\}$ crossed with $\kappa^{(0)} \in \{1, 2.5, 5, 10, 15, 20\}$, giving 54 fits per run, and retain the fit with the highest log-likelihood. For each model run, we define the \emph{interference} index as
\[
\mathrm{Interference} = 1 - \pi_A^\star,
\]
where $\pi_A^\star$ is the estimated weight on the task-A component at the chosen optimum. Values closer to 1 indicate stronger shift of behavior toward the task-B rule at retest.

\paragraph{Uncertainty estimates.}
Uncertainty in behavioral and representational plots is reported by panel type. For the per-run summary statistics shown in the transfer, interference, effective-dimensionality, and principal-angle panels, we use bootstrap 95\% confidence intervals: simulation runs are resampled with replacement $B = 1{,}000$ times within each (architecture, $\gamma$, similarity) cell, the summary statistic of interest is recomputed on each bootstrap sample, and we report the point estimate together with the central 95\% interval (2.5th to 97.5th percentile) as error bar. For the smoothed accuracy and loss curves (Figures showing the $A1 \rightarrow B \rightarrow A2$ time course), we instead report the participant-level mean $\pm$ SEM at each trial, with both the mean and the SEM passed through a centred rolling-mean window of 25 trials. All hypothesis tests reported in the main text are descriptive and based on these distributions; we do not perform formal multiple-comparison corrections.

\paragraph{Effective dimensionality and principal angles.}
For each model run, architecture, similarity, and phase $\phi \in \{A1, B, A2\}$, we form a per-phase data matrix $H_\phi \in \mathbb{R}^{12 \times h}$ by stacking the final-time-step hidden states for the 12 canonical-sweep stimuli (6 task-A stimuli followed by 6 task-B stimuli) of that phase, where $h$ is the total hidden width. PCA is fit on $H_\phi$ alone, and the effective dimensionality at variance threshold $\tau$ is
\[
d_{\mathrm{eff}}(\tau) = \min\!\left\{ k \;:\; \frac{\sum_{j=1}^k \lambda_j}{\sum_{j=1}^{D} \lambda_j} \geq \tau \right\},
\]
where $\{\lambda_j\}$ are the eigenvalues of the centred covariance of $H_\phi$ sorted in decreasing order and $D = \min(12, h)$. We report $d_{\mathrm{eff}}(0.99)$ at $\phi = B$ in the corresponding figure panels. We verified that the qualitative patterns reported in the Results are robust to varying the threshold between 95\% and 99\%.

To compute principal angles between task-specific subspaces, we work from the post-$B$ matrix $H_B \in \mathbb{R}^{12 \times h}$ defined above and split it by task identity along the canonical stimulus sweep, yielding $H^{(A)} \in \mathbb{R}^{6 \times h}$ (first 6 rows: task A) and $H^{(B)} \in \mathbb{R}^{6 \times h}$ (last 6 rows: task B). We then fit PCA separately on $H^{(A)}$ and $H^{(B)}$ and retain the first two principal components of each, yielding two-dimensional subspaces $\mathcal{S}_A$ and $\mathcal{S}_B$ in the shared hidden space. Let $U_A, U_B \in \mathbb{R}^{h \times 2}$ be orthonormal bases of $\mathcal{S}_A, \mathcal{S}_B$, and let $\sigma_1 \geq \sigma_2$ be the singular values of $U_A^\top U_B$. We report the first (smallest) principal angle in degrees,
\[
\theta_1 = \arccos(\sigma_1),
\]
as our scalar summary. The first principal angle ranges from $0$ (when $\mathcal{S}_A$ and $\mathcal{S}_B$ share at least one direction) to $90^\circ$ (when $\mathcal{S}_A$ and $\mathcal{S}_B$ are mutually orthogonal in the ambient space); larger values therefore correspond to stronger separation of task-specific subspaces.

\paragraph{PCA visualizations.}
For the qualitative 3D PCA trajectories shown in the main figures, we fit a single PCA on hidden states concatenated across phases $A1$, $B$, and $A2$ for a representative run at each architecture and initialization regime, using the same canonical stimulus sweep as above. We then project the hidden states for each phase and task onto the first three principal components and plot the resulting trajectories as closed loops connecting stimuli in sweep order. Colors encode phase and task identity, as described in the figure captions. These visualizations are used solely for illustration and do not enter into any quantitative analysis.


\section{Appendix - Ablation Studies}
\label{sec:appendix:ablation-studies}

The ablation experiments test whether our main findings can be explained by simple capacity differences or idiosyncrasies of a particular architecture, rather than by an interaction between representational dimensionality and modular structure. Across variations in module width, input routing, inter-module connectivity, initialization scope, and recurrent depth, we consistently observed that (i) modular architectures and single baselines behave similarly in high-dimensional (large-$\gamma$) regimes, and (ii) in low-dimensional (small-$\gamma$) regimes, modular networks retain their selective advantage: reduced interference and a graded, similarity–dependent organization of task subspaces. These results therefore reinforce the central claims of the paper by showing that the conditional benefit of modularity and the graded similarity–alignment pattern are robust to architectural and initialization choices and cannot be attributed to trivial changes in capacity or depth.

All ablation experiments reused the main protocol, including participant-derived schedules, the $A1 \rightarrow B \rightarrow A2$ training sequence, SGD optimization, $\gamma$-scaling, and all behavioral and representational metrics (accuracy, transfer, interference, effective dimensionality, and principal angles). Each ablation varied a single architectural or initialization factor while keeping all other settings fixed.

\subsection{A.5.1 Module-Size Sweep}

\paragraph{Question.}
Does the conditional benefit of modularity depend on network width or total recurrent capacity?

\paragraph{Setup.}
We evaluated the main task-routed modular architecture at four per-module hidden sizes $h_m$.
For each modular configuration, we constructed a single-network baseline with approximately matched total recurrent state width $2h_m$.
Table~\ref{tab:ablation:module-sizes} summarizes the configurations.

\begin{center}
\begin{tabular}{lccc}
\toprule
$h_m$ (modular) & Total modular state & Single hidden size & Notes \\
\midrule
6  & 12 & 12  & width-matched \\
12 & 24 & 25  & closest available single width \\
25 & 50 & 50  & main configuration (state-matched) \\
50 & 100 & 100 & widened configuration \\
\bottomrule
\end{tabular}
\end{center}
\label{tab:ablation:module-sizes}

\paragraph{Outcome.}
Across all widths, we observed the same qualitative pattern: in high-dimensional (large-$\gamma$) regimes, modular and single architectures behaved similarly, whereas in low-dimensional (small-$\gamma$) regimes the modular network showed reduced interference and a clearer graded dependence of subspace alignment on task similarity. Increasing width did not eliminate the graded similarity–alignment pattern or the conditional benefit of modularity, arguing against a simple capacity-based explanation.

\subsection{A.5.2 Shared-Input Modular Ablation}

\paragraph{Question.}
Are modular benefits driven specifically by task-dependent input routing, or does block-diagonal recurrence alone suffice?

\paragraph{Setup.}
We compared the main task-routed modular architecture to a shared-input modular variant.
Both variants had:
(i) two recurrent modules of identical size,
(ii) block-diagonal recurrent connectivity with no inter-module recurrent communication,
and (iii) a shared linear readout.
They differed only in the input routing scheme:

\begin{center}
\begin{tabular}{lcc}
\toprule
Property & Task-routed modular & Shared-input modular \\
\midrule
Input delivery & Input delivered only to task-relevant module & Full input delivered to both modules \\
Recurrent structure & Block-diagonal ($W_A$, $W_B$) & Block-diagonal ($W_A$, $W_B$) \\
Readout & Shared linear readout & Shared linear readout \\
\bottomrule
\end{tabular}
\end{center}

\paragraph{Outcome.}
Task-routed modular networks reproduced the main result: in low-dimensional regimes they showed robust reductions in interference and a graded similarity–dependent organization of task subspaces. The shared-input modular variant exhibited weaker improvements, with more entangled task subspaces and less pronounced similarity-dependent angles, indicating that task-dependent input gating contributes to the observed modular advantages beyond recurrent structural separation alone.

\subsection{A.5.3 Inter-Module Sparsity Sweep}

\paragraph{Question.}
Is strict module isolation necessary, or do the effects persist when modules are partially recurrently coupled?

\paragraph{Setup.}
We introduced sparse off-diagonal recurrent connectivity between modules in the modular architecture.
For the $h_m = 25$ configuration, we varied the density of inter-module connections
\[
s \in \{0.0,\, 0.3,\, 0.5,\, 0.7,\, 0.9,\, 1.0\},
\]
where $s=0$ corresponds to the main block-diagonal condition and $s=1$ to fully dense inter-module communication.
This sweep was performed for both task-routed and shared-input modular networks.

\paragraph{Outcome.}
Moderate levels of inter-module connectivity ($s \leq 0.5$) preserved the qualitative pattern seen in the strictly isolated case: modular architectures in low-dimensional regimes continued to show reduced interference and graded similarity–dependent subspace organization. At very high inter-module densities ($s \approx 1.0$), task subspaces became more entangled and the similarity gradient in principal angles was attenuated, consistent with a breakdown of effective modular separation. These results suggest that partial, but not total, isolation is sufficient for the main modular advantages to emerge.

\subsection{A.5.4 Initialization-Scope Ablation}

\paragraph{Question.}
Do rich versus lazy regimes arise primarily from scaling the recurrent dynamics, the input pathway, or both?

\paragraph{Setup.}
In the main experiments, the initialization scale $\gamma$ was applied globally to all trainable recurrent parameters after default initialization.
To separate the contributions of input scaling and recurrent-state scaling, we compared two initialization scopes for the $h_m = 25$ modular architecture across multiple sparsity settings:

\begin{center}
\begin{tabular}{lcc}
\toprule
Scope & Parameters scaled by $\gamma$ & Description \\
\midrule
Global (main) & Input and recurrent weights & Full recurrent-regime manipulation \\
Input-only & Input weights only & Input gain manipulation \\
\bottomrule
\end{tabular}
\end{center}

\paragraph{Outcome.}
Global scaling produced the pronounced transition between high-dimensional and low-dimensional regimes reported in the main text, along with the associated changes in transfer, interference, and geometry. Input-only scaling induced smaller changes in effective dimensionality and weaker modulation of the behavioral and geometric measures. This pattern supports our interpretation of $\gamma$ as primarily controlling the recurrent representational regime, rather than acting as a trivial rescaling of input gain.

\subsection{A.5.5 Depth Ablation}

\paragraph{Question.}
Are the observed modular advantages specific to shallow recurrent networks?

\paragraph{Setup.}
To examine the effect of recurrent depth, we varied the number of recurrent layers while keeping hidden-state width fixed.
For both single and modular architectures, we evaluated recurrent depths
\[
L \in \{1,\, 2,\, 3\}.
\]
All other training and analysis settings were identical to the main experiments.

\paragraph{Outcome.}
Increasing recurrent depth up to $L=3$ did not qualitatively change the main findings: in high-dimensional regimes, modular and single architectures behaved similarly, while in low-dimensional regimes the modular networks continued to show reduced interference and a clearer graded dependence of subspace alignment on task similarity. This indicates that the conditional benefit of modularity is not tied to a specific recurrent depth, but instead reflects how architectural separation interacts with representational dimensionality.

\end{document}